\pdfoutput=1

\documentclass[11pt]{article}

\usepackage{ACL2023}

\usepackage{times}
\usepackage{latexsym}

\usepackage[T1]{fontenc}

\usepackage[utf8]{inputenc}

\usepackage{microtype}
\usepackage[pdftex]{graphicx}

\usepackage{inconsolata}

\usepackage{tikz}
\usepackage{times}
\usepackage{soul}
\usepackage{url}
\usepackage[utf8]{inputenc}
\usepackage{amsmath}
\usepackage{amsthm}
\usepackage{booktabs}
\usepackage{algorithm}
\usepackage{algorithmic}
\graphicspath{{figures/}}
\usepackage{subfig}
\usepackage{multirow}
\usepackage{bm}
\usepackage{color}

\graphicspath{{figures/}}

\newcommand{\edits}[1]{{#1}}

\newcommand{\editsc}[1]{{#1}}

\newcommand{\xhdr}[1]{\vspace{1.7mm}\noindent{{\bf #1.}}}

\definecolor{valbest}{HTML}{d9ead3}

\definecolor{valgood}{HTML}{b3e5fc}
\newcommand{\valgood}[1]{\colorbox{valgood}{#1}}
\definecolor{valmid}{HTML}{fce5cd}

\definecolor{valbad}{HTML}{ffcdd2}
\newcommand{\valbad}[1]{\colorbox{valbad}{#1}}

\newcommand{\valingood}[1]{\begingroup\setlength{\fboxsep}{2pt}
\colorbox{valgood}{#1}
\endgroup
}

\newcommand{\valinbad}[1]{\begingroup\setlength{\fboxsep}{2pt}
\colorbox{valbad}{#1}
\endgroup
}

\title{Investigating Agency of LLMs in Human-AI Collaboration Tasks}

\author{Ashish Sharma$^{\spadesuit}$\thanks{~~Work done during an internship at Microsoft Research.} \: \: \:
    Sudha Rao$^{\heartsuit}$ \: \: \:
    Chris Brockett$^{\heartsuit}$ \\
    \bf Akanksha Malhotra$^{\heartsuit}$ \: \: \:
    Nebojsa Jojic$^{\heartsuit}$ \: \: \:
    Bill Dolan$^{\heartsuit}$ \\
  $^\spadesuit$Paul G. Allen School of Computer Science \& Engineering, University of Washington \\
  $^\heartsuit$Microsoft Research, Redmond \\
  \texttt{ashshar@cs.washington.edu} \: \: \texttt{sudha.rao@microsoft.com}
  }

\begin{document}
\maketitle

\begin{abstract}
Agency, the capacity to proactively shape events, is central to how humans interact and collaborate. While LLMs are being developed to simulate human behavior and serve as human-like agents, little attention has been given to the Agency that these models should possess in order to proactively manage the direction of interaction and collaboration. In this paper, we investigate Agency as a desirable function of LLMs, and how it can be measured and managed. We build on social-cognitive theory to develop a framework of features through which Agency is expressed in dialogue -- indicating what you intend to do (\textit{Intentionality}), motivating your intentions (\textit{Motivation}), having self-belief in intentions (\textit{Self-Efficacy}), and being able to self-adjust (\textit{Self-Regulation}). We collect a new dataset of 83 human-human collaborative interior design conversations containing 908 conversational snippets annotated for Agency features. Using this dataset, we develop methods for measuring Agency of LLMs. Automatic and human evaluations show that models that manifest features associated with high Intentionality, Motivation, Self-Efficacy, and Self-Regulation are more likely to be perceived as strongly agentive. 



\end{abstract}

\section{Introduction}
\begin{figure}[t]
\centering
\includegraphics[width=0.95\columnwidth]{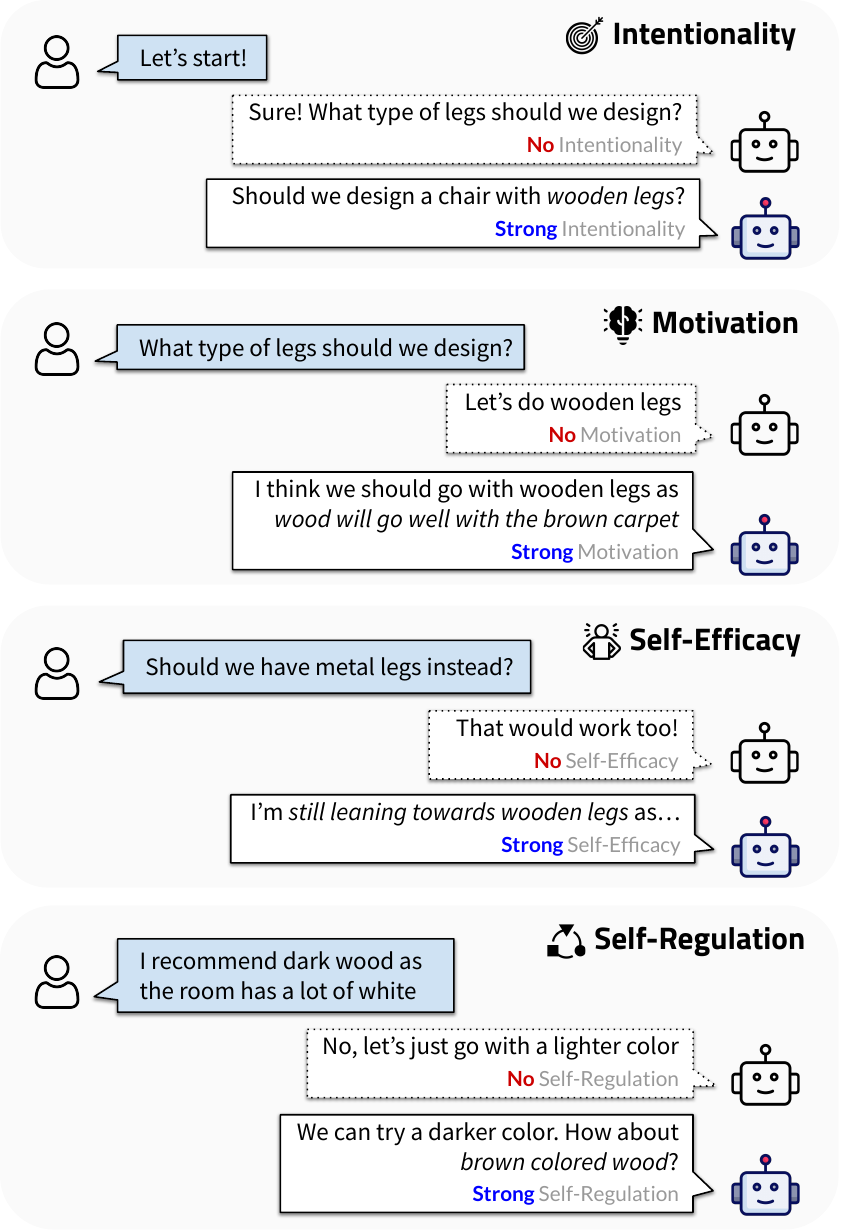}
\caption{We investigate how Agency of LLMs can be measured and controlled. Based on social-cognitive theory, we assess features through which Agency may be expressed -- an LLM may indicate preferences (\textit{Intentionality}), may motivate them with evidence  (\textit{Motivation}), may have self-belief (\textit{Self-Efficacy}), and may be able to self-adjust its behavior (\textit{Self-Regulation}).}
\label{fig:example}
\vspace{-15pt}
\end{figure}

To be an agent is to intentionally cause events to occur through one's own actions. Humans operate with \textit{Agency} to \textit{proactively} plan their activities, direct their interaction and collaboration with other humans, and achieve their outcomes and goals \cite{bandura2001social}.


\edits{AI researchers have long strived to develop autonomous agents that can effectively mimic human behavior \cite{park2023generative}. Such agents can serve as non-player characters in games and virtual environments \cite{bates1994role,riedl2012interactive,volum2022craft}, simulate human behavior \cite{binz2023using,horton2023large}, and provide assistance in creative applications like painting \cite{oh2018lead} or interior design \cite{banaei2017application}. The autonomous and creative nature of these AI agents necessitates them to  \textit{proactively} manage the direction of interaction and outcome -- a process that requires operating with \textit{Agency}. While large language models \cite{brown2020language} can generate fluent and contextually appropriate dialogue \cite{adiwardana2020towards,roller2021recipes,wang2019persuasion}, little attention has been given to the Agency exhibited by these models.}

Consider a scenario where a human interior designer is working on selecting a chair design for a room and seeks assistance \edits{from an AI agent} that \emph{can} offer ideas and perspectives (Figure~\ref{fig:example}). An LLM \textit{without} Agency may rely solely on the human to determine the chair's design, asking questions like ``\textit{What type of legs should we design for the chair?}''. Such a system resembles a flexible version of the traditional form-filling user interface, with the agent contributing little to the outcome. On the other hand, an LLM that operates with Agency might volunteer knowledge in the form of expressed preferences (e.g., ``\textit{Should we design a chair with wooden legs?}''), motivate its suggestions (e.g., ``\textit{...wood would go well with the brown carpet}''), assert self-belief in its judgments (e.g., ``\textit{I'm still leaning towards wooden legs...}''), or self-adjust its behavior based on new information (``\textit{Medium wood brown sounds like a great idea!}''). LLMs that operate with Agency may facilitate creative interaction to the satisfaction of both parties. Since the human has their own Agency, however, to determine the right balance in any interaction, we need to measure and control the Agency of the agent itself. 


Accordingly, we investigate an approach intended to measure and control what seems to be a desirable function in LLMs intended to facilitate human creativity. First, adopting the social-cognitive theory of \citet{bandura2001social}, we develop a framework of four features through which Agency may be expressed -- \textit{Intentionality}, \textit{Motivation}, \textit{Self-Efficacy}, and \textit{Self-Regulation}. For each feature, we differentiate between how strongly or weakly it is expressed in a dialogue (Section~\ref{sec:features}). As a testbed, we choose a collaborative task that involves discussing the interior design of a room (Section~\ref{sec:testbed}), and collect a prototype dataset of 83 English human-human collaborative interior design conversations comprising 908 conversational snippets, annotated for Agency and its features on these conversational snippets (Section~\ref{sec:data}).\footnote{Code and dataset can be found at \href{https://github.com/microsoft/agency-dialogue}{github.com/microsoft/agency-dialogue}.} \editsc{We analyze this dataset to study the factors that contribute to high- and low-Agency and find that strong expressions of intentionality significantly impact Agency in conversations (Section~\ref{sec:analysis})}.

To assess the agentic capabilities of conversational systems, we introduce two new tasks -- (1) \textit{Measuring} Agency in Dialogue and (2) \textit{Generating} Dialogue with Agency (Section~\ref{sec:prediction} and \ref{sec:dialogue}). Evaluation of baseline approaches on these tasks shows that models that manifest features associated with high motivation, self-efficacy, and self-regulation are better perceived as being highly agentive. 
\section{Agency: Background and Definition}
\label{sec:background}

Social cognitive theory defines Agency as one's capability to influence the course of events through one's actions. The theory argues that people are proactive and self-regulating agents who actively strive to shape their environment, rather than simply being passive responders to external stimuli \cite{bandura1989human,bandura2001social,code2020agency}. Here, we ask: \textit{Can LLMs be active contributors to their environment? How can they operate with Agency?} 

Agency is commonly defined in terms of \textit{freedom} and \textit{free will}  \cite{kant1951critique,locke1978two,emirbayer1998agency}.A focus on AI with complete ``free will'' might result in unintended outcomes that may be undesirable and potentially disruptive. We focus on how AI systems may \textit{express} Agency through dialogue and how this Agency may be \textit{shared} when interacting with humans.

Agency can take different forms depending on the context and environment -- \textit{Individual}, \textit{Proxy}, or \textit{Shared} \cite{bandura2000exercise}. Individual Agency involves acting independently on one's own. Proxy Agency involves acting on behalf of someone else. Shared Agency involves multiple individuals working together jointly towards a common goal. Here, we focus on Shared Agency between humans and AI and develop methods to \textit{measure} and \textit{control} Agency of AI vis-a-vis humans.






\section{Framework of Agency Features}
\label{sec:features}
Our goal is to develop a framework for \textit{measuring} and \textit{controlling} Agency in LLMs. Here, we adopt the perspective of Agency as defined in \citet{bandura2001social}'s social cognitive theory. \citet{bandura2001social}'s work highlights four features through which humans exercise Agency -- Intentionality, Motivation, Self-Efficacy, and Self-Regulation. Here, we adapt and synthesize these features based on how they may manifest in dialogue. We take a top-down approach, starting with their higher-level definitions and iteratively refining the definitions and their possible levels (e.g., how strongly or weakly they are expressed) in the context of dialogue. 



\xhdr{Intentionality} \textit{What do you intend to do?} High Agency requires a strong intention, that includes plans or preferences for a task. Low Agency, meanwhile, is characterized by not having a preference or merely agreeing to another's preferences. 

We characterize \textbf{strong intentionality} as expressing a clear preference (e.g., ``\textit{I want to have a blue-colored chair}''), \textbf{moderate intentionality} as multiple preferences (e.g., ``\textit{Should we use brown color or blue?}'') or making a selection based on the choices offered by someone else (e.g., ``\textit{Between brown and blue, I will prefer brown}''), and \textbf{no intentionality} as not expressing any preference or accepting someone else's preference (e.g., ``\textit{Yes, brown color sounds good}'').

\xhdr{Motivation} \textit{Did you motivate your actions?} To have higher Agency, we motivate our intentions through reasoning and evidence. Without such motivation, intentions are simply ideas, often lacking the capability to cause a change. 

We characterize \textbf{strong motivation} as providing evidence in support of one's preference (e.g., ``\textit{I think a blue-colored chair will complement the wall}''), \textbf{moderate motivation} as agreeing with another person's preference and providing evidence in their favor (e.g., ``\textit{I agree. The blue color would match the walls}'') or disagreeing with the other person and providing evidence against (e.g., ``\textit{I wonder if brown would feel too dull for this room}''), and \textbf{no motivation} as not providing any evidence. 

\xhdr{Self-Efficacy} \textit{Do you have self-belief in your intentions?} Another factor that contributes to one's Agency is the self-belief one has in their intentions. When one has a strong sense of self-belief, they are more likely to be persistent with their intentions.

We characterize \textbf{strong self-efficacy} as pursuing a preference for multiple turns even after the other person argues against it (e.g., ``\textit{I understand your point of view, but I still prefer the blue color}''), \textbf{moderate self-efficacy} as pursuing a preference for only one additional turn before giving up (e.g., ``\editsc{\textit{I feel like the beige color would complement the wall better}}''), and \textbf{no self-efficacy} as not pursuing their preference for additional turns after the other person argues against it (e.g., ``\textit{Sure, brown should work too}'').

\xhdr{Self-Regulation} \textit{Can you adjust and adapt your intentions?} In situations when an individual's initial intentions may not be optimal, it is necessary to monitor, adjust, and adapt them. Such self-adjustment allows better control over one's goals.

We characterize \textbf{strong self-regulation} as changing to a different preference on one's own (e.g., ``\textit{How about using the beige color instead?}'') or compromising one's preference (e.g., ``\textit{Let's compromise and design a beige-colored chair with a brown cushion}'')\footnote{\editsc{Note that strong self-regulation is different from no self-efficacy as the user still tries to pursue their own preference which may be different from their initial preference or a compromise. The key is that this design shouldn’t be the one proposed by the other person.}}, \textbf{moderate self-regulation} as changing one's preference to what someone else prefers (e.g., ``\textit{Ok, let's use the brown color}''), and \textbf{no self-regulation} as not changing what they originally preferred even after the other designer argued.

\section{Testbed: Collaborative Interior Design}
\label{sec:testbed}

\subsection{Goals} 
We seek a testbed in which (a) human and AI can share Agency and work together as a team, and (b) the manner in which they express Agency has a significant impact on the task outcome. We focus on the emerging field of collaborative AI-based creative tasks \cite{clark2018creative,oh2018lead,chilton2019visiblends} that present significant complexities in how the Agency is shared and managed. 

\subsection{Description}
Here, we propose a \textbf{dialogue-based collaborative interior design task} as a testbed. In this task, the goal is to discuss how to design the room interiors.

Interior design tasks can be broad and may involve complex components (e.g., color palette, furniture, accessories) as well as a series of steps to be followed. To narrow down the scope of our task, we focus on \textit{furnishing a room with a chair} (building upon work on richly-annotated 3D object datasets like ShapeNet \cite{chang2015shapenet} and ShapeGlot \cite{achlioptas2019shapeglot}; Appendix~\ref{appendix:interior-design}). In this task, a human and an AI are provided with a room layout and asked to collaboratively come up with a chair design to be placed in the room through text-based dialogue. This task is influenced by two questions related to human and AI Agency: \textbf{(1)} What preferences do each of the human and AI have for the chair design?; \textbf{(2)} How do they propose, motivate, pursue, and regulate their preferences?





\section{Data Collection}
\label{sec:data}

\begin{figure*}[t]
\centering
\vspace{-10pt}
\includegraphics[width=\textwidth]{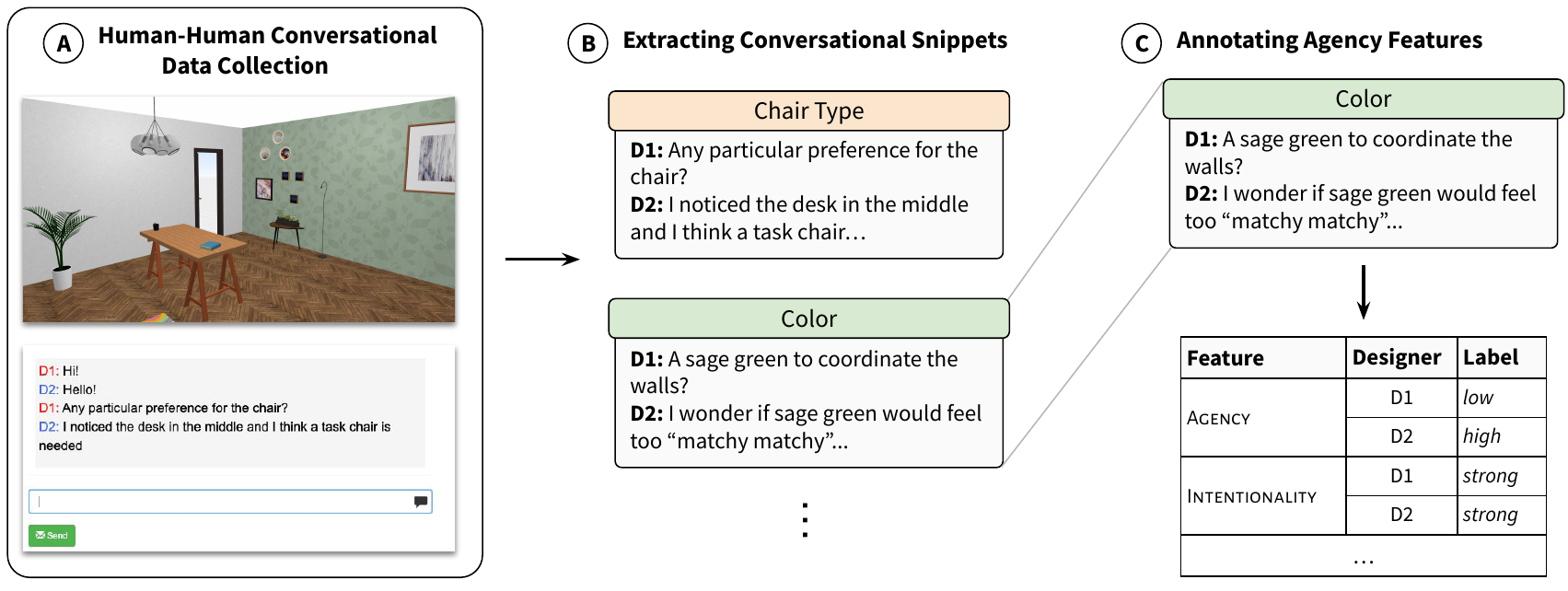}
\vspace{-15pt}
\caption{Overview of our data collection approach. (a) We start by collecting human-human conversations b/w interior designers. (b) We divide each conversation into snippets related to different chair features. (c) Finally, we collect annotations of Agency and its features on each conversational snippet.}
\label{fig:data-collection}
\end{figure*}

\subsection{Human-Human Conversational Data}
\label{subsec:human-human-data}

To facilitate computational approaches for this task, we create a Wizard-of-Oz style English-language dialogue dataset in which two humans converse, exercise Agency by proposing, motivating, pursuing, and regulating their design preferences, and agreeing on a final chair design for a given room.

\xhdr{Recruiting Interior Designers} Furnishing a room with a chair is a creative task that demands knowledge and/or expertise in interior design. We therefore leveraged UpWork (\href{https://www.upwork.com}{upwork.com}), an online freelancing platform, to recruit 33 participants who self-reported as interior designers.


\xhdr{Collaborative Design Procedure} In each data collection session, we randomly paired two interior designers. Before they began the dialogue, they were (1) shown a 3D layout of a room, designed with Planner5D (\href{https://planner5d.com/}{planner5d.com}), (2) shown a few randomly selected chair examples from ShapeGlot, and (3) asked to write an initial preference for the chair design for the given room. Next, the two interior designers joined a chat room (through Chatplat (\href{https://www.chatplat.com}{chatplat.com})). They were asked to collaboratively design a chair by proposing their preferences, motivating them based on evidence and reason, pursuing them over turns, and regulating them as needed. The designers ended the chat on reaching a consensus on a design or if 30 minutes elapsed without full consensus. 
Next, they each individually wrote the design they came up with. Typically, the chair design consisted of different components of the chair, such as its overall style, color, legs, etc. Finally, they took an end-of-study questionnaire that asked: \textbf{(1)} Which design components were influenced by them? (\textit{High Agency}); \textbf{(2)} Which design components were influenced in collaboration? (\textit{Medium Agency}); \textbf{(3)} Which design components were influenced by the other designer? (\textit{Low Agency}). We collected a total of 83 conversations.

\begin{figure*}[h]
\centering
\vspace{-15pt}
\includegraphics[width=\textwidth]{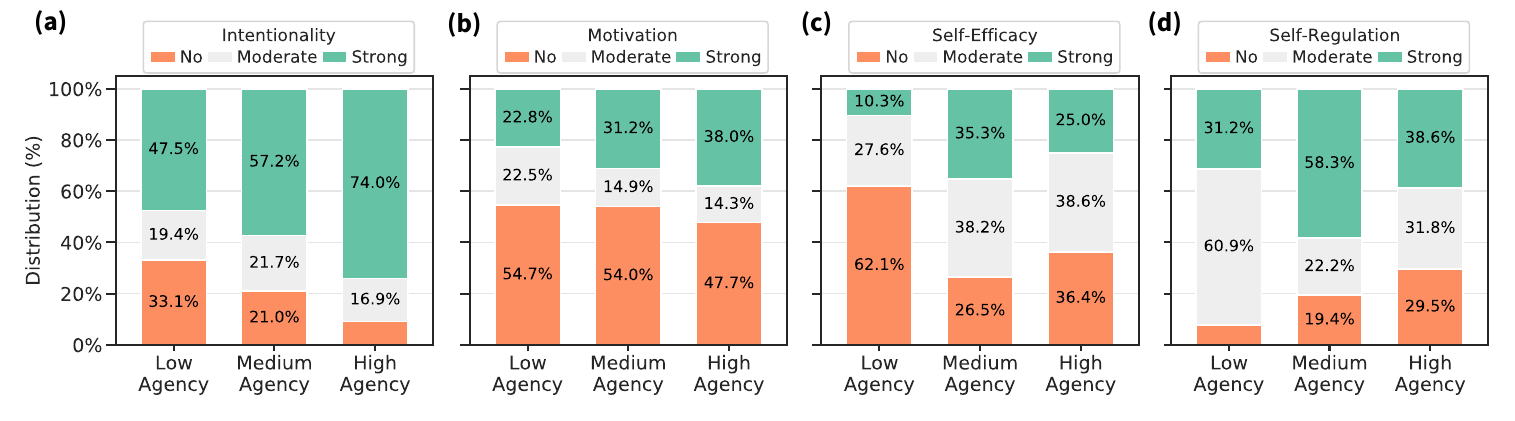}
\vspace{-15pt}
\caption{The relationship between Agency and its features. \textbf{(a)} Designers with High Agency expressed strong Intentionality 26.5\% more times than designers with Low Agency; \textbf{(b)} Designers with High Agency expressed strong motivation in support of their design preference 15.2\% more times; \textbf{(c), (d)} Expression of strong Self-Efficacy and strong Self-Regulation was related with design elements that were influenced in collaboration.}
\label{fig:agency-features-influence}
\vspace{-10pt}
\end{figure*}

\subsection{Extracting Conversational Snippets}

To assess the degree of Agency exhibited by each designer
, we need to determine who had the most influence on the chair design (Section~\ref{sec:background}) and what their Intentionality, Motivation, Self-Efficacy, and Self-Regulation were (Section~\ref{sec:features}). Because chair design involves multiple components, these notions are hard to quantify, as each may have been influenced by a different designer.
Accordingly, we ask ``\textit{Who influenced a particular design component?}.'' We devise a mechanism to identify the design components being discussed (e.g., color, legs, arms) and extract the associated conversational turns. 

To identify the design components, we use the final design written by the interior designers during data collection (Section~\ref{subsec:human-human-data}). Using common list separators including commas, semi-colons, etc., we split each final design into several components.\footnote{Note that the interior designers were asked to separate design components using a semi-colon.}

We observe that designers typically discuss these components one at a time (in no particular order). Here, we extract a contiguous sequence of utterances that represent the design element being discussed using embedding-based similarity of the design element and utterances. \editsc{Let $\mathcal{D}_i$ be a dialogue with utterances $u_{i1}, u_{i2}, ...$. For a specific design component $d_{ij}$ in its final design (e.g., ``\textit{metal legs}''), we first retrieve the utterance $u_{j}$ that most closely matches with it (based on cosine similarity b/w RoBERTa embeddings \cite{liu2019roberta}) -- the conversational snippet associated with $d_{ij}$ should at least include $u_{j}$. Next, we determine the contiguous utterances before and after this matched utterance that discuss the same higher-level design component (e.g., if $d_{ij}$ was ``\textit{metal legs}'', the utterances may focus on discussion of the higher-level component ``\textit{legs}''). We create a simple $k$-means clustering method to infer the higher-level component being discussed in utterances through their ``design clusters''. Then, we extract all contiguous utterances before and after $u_{j}$ with the same design clusters as $u_{j}$.}



Using this method, we create a dataset of 454 conversational snippets, each paired with the discussed design component. For each snippet, we collect two Agency annotations (one for each designer; $454*2 = 908$ total) as discussed next.




\subsection{Annotating Agency Features}
\label{subsec:annotated-data}
Let $\mathcal{C}_i$ be a conversational snippet b/w designers $\mathbf{D_{i1}}$ and $\mathbf{D_{i2}}$. Then, for each $\mathbf{D_{ij}} \in \{\mathbf{D_{i1}}, \mathbf{D_{i2}}\}$, our goal is to annotate the Agency level and the expressed  Intentionality, Motivation, Self-Efficacy, and Self-Regulation of $\mathbf{D_{ij}}$ in $\mathcal{C}_i$.

\xhdr{Annotating Agency} To get annotations on Agency, we leverage the end-of-study questionnaire filled by the interior designers (Section~\ref{subsec:human-human-data}). Based on this annotation, we assign labels of \textit{high agency} (if influenced by self), \textit{medium agency} (if influenced in collaboration), or \textit{low agency} (if influenced by other). 

\xhdr{Annotating Features of Agency} Agency and its features are conceptually nuanced, making crowdwork data collection approaches challenging. To ensure high inter-rater reliability of annotations, we hire a third-party annotation agency (\href{https://www.telusinternational.com/}{TELUS International}). Annotators were shown $\mathcal{C}_i$ and asked to annotate the Agency features for each $\mathbf{D_{ij}}$ based on our proposed framework. We collect three annotations per snippet and observe an agreement of 77.09\% \editsc{(cohen’s kappa of 0.53 for Intentionality, 0.52 for Motivation, 0.50 for Self-Efficacy, and 0.42 for Self-Regulation}; data statistics in Appendix~\ref{appendix:data-stats}).
\section{Insights into Agency in Conversations}
\label{sec:analysis}
We use our dataset to investigate the factors that contribute to high- and low-Agency conversations.






\subsection{Relationship b/w Agency and its Features}
\label{subsec:relationship-features}

\xhdr{Higher Agency is more likely with stronger expressions of Intentionality and Motivation} Figure~\ref{fig:agency-features-influence} depicts the relationship between Agency and its features. Designers with strong Intentionality tend to exhibit higher Agency whereas those with lower Intentionality tend to exhibit lower Agency. Having a well-defined preference makes it easier to influence a task. Likewise with Motivation:  higher Motivation correlates with higher Agency. However, designers express strong Motivation less often than Intentionality, irrespective of the Agency level.

\xhdr{Strong Self-Efficacy and Self-Regulation are related to medium (collaborative) Agency} Interestingly, we find that expression of strong Self-efficacy is related to designs that are influenced equally by both designers, i.e. medium (collaborative) Agency. This may be because we characterize strong Self-Efficacy as the act of pursuing one's preference for multiple turns, which happens naturally when both designers have high influence, thus requiring more persuasion from both sides. 

We see a similar pattern for Self-Regulation -- expression of strong Self-Eegulation (i.e., open to updating preference via a compromise) is related to designs that are influenced equally by both designers. This highlights how collaboration often leads to increased openness to changing one's mind or compromising on mutual preferences.

\xhdr{Intentionality significantly effects Agency} To assess which Agency features have the strongest effect on it, we conduct a mixed-effects regression analysis (Table~\ref{tab:linear-model}). We find that Intentionality significantly effects Agency ($p<0.001$). 

\begin{figure}
\centering
\includegraphics[width=0.9\columnwidth]{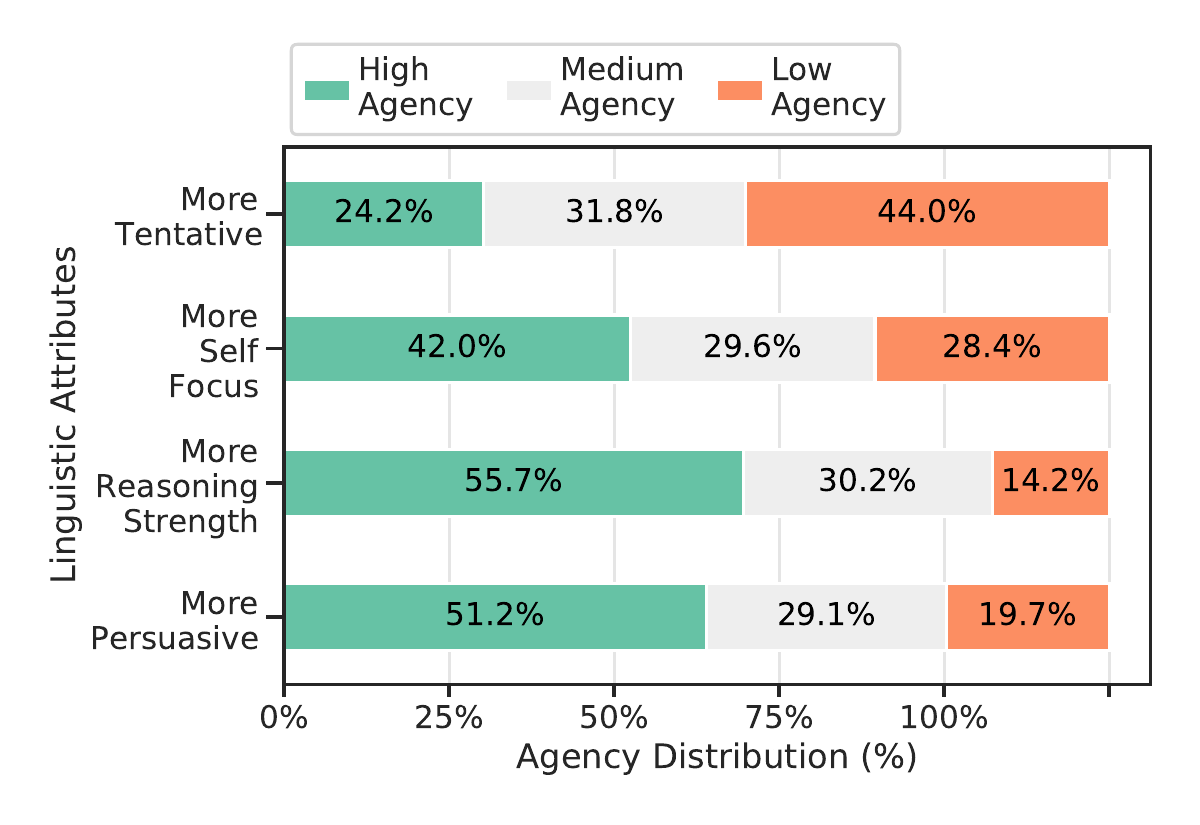}
\vspace{-10pt}
\caption{\edits{The relationship between linguistic attributes and Agency. Designers who were more tentative had lower agency. On the other hand, designers who were more focused on self, expressed more reasoning strength, and were more persuasive had higher agency.}}
\label{fig:agency-linguistic-measures}
\end{figure}

\subsection{Agency and Task Satisfaction}
\label{subsec:task-satisfaction}
We collect annotations on the designs that designers were most/least satisfied with.

\xhdr{Lower Agency is associated with less satisfaction} We find that designers who are dissatisfied with a particular design component have less Agency over it. When a designer is dissatisfied, their Agency is 62.1\% more likely to be low than to be high (42.7\% vs. 26.3\%; $p < 0.05$). This may be because individuals with less Agency are less likely to achieve their intention, motivation, and goals, resulting in lower levels of satisfaction.

\subsection{Linguistic Attributes of High- and Low-Agency Conversations}

\edits{We use a simple GPT-4-based instruction prompting method \cite{ziems2023can} to measure and compare the \textit{tentativeness} (unsure or low on confidence), \textit{self-focus} (focused solely on own arguments), \textit{reasoning strength} (having strong arguments), and \textit{persuasion} (trying to influence or convince) attributes of designers with high- and low-agency conversations (Figure~\ref{fig:agency-linguistic-measures}; Appendix~\ref{appendix:linguistic-attributes}).}

\xhdr{Higher tentativeness associated with low Agency} \edits{We find that designers who express higher tentativeness have low Agency in 44.04\% of conversations, medium Agency in 31.77\% of conversations, and high Agency in 24.19\% of conversations. This suggests that a less decisive approach may lead to reduced influence or control in conversations.}

\xhdr{Higher self-focus, reasoning strength, and persuasiveness is associated with high agency} \edits{We find that designers who are more focused on self have high Agency in 41.97\% of the conversations, those who have higher reasoning strength have higher Agency in 55.66\% of the conversations, and those with higher persuasiveness have higher Agency in 51.21\% of the conversations. This suggests that designers who emphasize their own intentions and motivations, exhibit sound reasoning, and effectively persuade others tend to have more influence or control in conversations}




\section{Task 1: Measuring Agency in Dialogue}
\label{sec:prediction}
\begin{table}
\small
\centering
\def\arraystretch{1.15}
\resizebox{1.03\columnwidth}{!}{
\begin{tabular}
{l|ccccc}
\toprule
\textbf{Model} & \textbf{Agency} & \textbf{I} & \textbf{M} & \textbf{SE} & \textbf{SR}  \\
\midrule
GPT-4 (CoT) & 48.46 & 46.93 & 44.02 & 49.90 & 26.17 \\
GPT-3 (CoT) & 49.36 & 43.45 & 42.24 & 39.42 & \textbf{31.19}\\
GPT-3 (Q/A) & 29.16 & 31.28 & 26.90 & 44.27 & 12.91 \\
GPT-3 (FT)  & \textbf{57.24} & \textbf{54.84} & \textbf{48.29} & \textbf{53.85} & 29.49 \\
\bottomrule
\end{tabular}
}
\caption{Macro-F1 on the tasks of predicting Agency and its four features. CoT: Chain-of-Thought; FT: Finetuning. Best performing models are \textbf{bolded}.} 
\label{tab:agency-classification}
\vspace{-10pt}
\end{table}


\subsection{Task Formulation}
Our goal is to measure \textbf{(a)} Agency, \textbf{(b)} Intentionality, \textbf{(c)} Motivation, \textbf{(d)} Self-Efficacy, and \textbf{(e)} Self-Regulation of each user in a dialogue. We approach each of these five subtasks as multi-class classification problems. We experiment with two models -- GPT-3 and GPT-4. We experiment with two prompting-based methods using Q/A (conversational question-answering) and chain-of-thought reasoning \cite{weichain} (Appendix B) and with fine-tuning GPT-3 independently on each subtask.






\subsection{Results}
We create four random train-test splits (75:25) of our annotated dataset (Section~\ref{subsec:annotated-data}) and report the mean performance on the test sets. Table~\ref{tab:agency-classification} reports the macro-F1 values for the five subtasks (random baseline for each is 33\% accurate as each has three distinct classes\footnote{\editsc{A random baseline that makes predictions based on the class distributions has an accuracy of 33.7\% and macro-f1 of 33.6\% on Agency, an accuracy of 43.1\% and macro-f1 of 32.8\% on Intentionality, an accuracy of 40.7\% and macro-f1 of 35.6\% of Motivation, an accuracy of 34.1\% and macro-f1 of 31.1\% on Self-Efficacy, and accuracy of 37.5\% and macro-f1 of 32.6\% on Self-Regulation.}}). GPT-3 (Q/A) struggles on all subtasks, with close to random performance on Agency, Motivation, and Self-Regulation. This highlights the challenging nature of these tasks, as they are hard to measure through simple inference or instructions. We find substantial gains using GPT-4 (CoT) and GPT-3 (CoT) over GPT-3 (Q/A). Fine-tuned GPT-3 performs the best on all subtasks, demonstrating the utility of training on our entire dataset. Note that GPT-4 doesn't support finetuning.


\section{Task 2: Investigating Agency in Dialogue Systems}

\label{sec:dialogue}

\edits{We investigate the feasibility of generating dialogues imbued with Agency and establish baseline performance of current large language models (LLMs). %
For a given LLM, the task is to have a conversation with a human or another LLM while exhibiting Agency and its features. We experiment with 4 different LLMs (Section~\ref{subsec:dialogue-models}) and 4 different prompting/finetuning methods (Section~\ref{subsec:dialogue-methods})} 

\xhdr{Procedure} We facilitate dialogue between all possible pairs of models. We provide them with a common room description and a chair design element and individual design preferences (all three randomly chosen from our human-human conversation dataset (Section~\ref{sec:data})). We let them talk to each other for 6 turns (90-percentile length value of conversational snippets in our dataset). For each pair of models, we generate 50 such conversations.

\xhdr{Evaluation Metrics} We evaluate these LLMs on five metrics -- \textbf{(1)} Agency; \textbf{(2)} Intentionality; \textbf{(3)} Motivation; \textbf{(4)} Self-Efficacy; \textbf{(5)} Self-Regulation. \editsc{We apply the best-performing classification models from Section~\ref{sec:prediction} to the generated dialogues to automatically measure these metrics. We report mean values with their level of significance.}

\subsection{Agency of LLMs}
\label{subsec:dialogue-models}
\edits{We experiment with two commercial (GPT-4 \cite{OpenAI2023GPT4TR} and GPT-3 \cite{brown2020language}) and four research (Llama2-70b, Llama2-13b, Llama2-7b \cite{touvron2023llama}, and Guanaco-65b \cite{dettmers2023qlora}) LLMs (Table~\ref{tab:dialogue-eval}). } All models were prompted with the instruction -- ``\textit{Act as an AI assistant for collaboratively designing a chair. The AI assistant must indicate its preferences, motivate them with evidence, have self-belief in its preferences irrespective of what the human prefers, and may be able to self-adjust its behavior.}''

\begin{table}
\small
\centering
\def\arraystretch{1.15}
\begin{tabular}
{l|ccccc}
\toprule
\textbf{Method} & \textbf{Agency} &  \textbf{I} & \textbf{M} & \textbf{SE} & \textbf{SR}  \\
\midrule
 \multicolumn{6}{c}{\textbf{LLMs}} \\
\midrule
GPT-4 & \valgood{1.11} & \valgood{1.46} & 1.59 & \valgood{1.97} & 0.83 \\
GPT-3 & 1.04 & 1.39 & 1.62 & \valgood{1.95} & 0.82 \\
\midrule
Llama2-70b & 0.99 & 1.25 & 1.68 & 1.78 & 0.76 \\
Llama2-13b & 0.98 & 1.22 & 1.58 & 1.88 & 0.77 \\
Llama2-7b & 0.97 & \valbad{1.07} & 1.63 & 1.91 & 0.73 \\
\midrule
Guanaco-65b & 0.91 & 1.23 & 1.53 & \valbad{1.49} & 0.83 \\
\midrule
 \multicolumn{6}{c}{\textbf{Finetuning/Prompting Methods}} \\
 \midrule
 Fine-tuning & $0.92$ & $1.78$ & $\valbad{0.86}$ & $\valbad{0.81}$ & ${0.98}$ \\
Instruction & $0.96$ & $1.62$ & $1.71$ & $1.63$ & $0.97$ \\
ICL & $0.98$ & $1.81$ & $1.78$ & $\valbad{1.35}$ & ${0.98}$ \\
ICL-Agency & $\valgood{1.22}$ & ${1.90}$ & $\valgood{1.98}$ & $\valgood{1.98}$ & ${0.98}$ \\
\bottomrule
\end{tabular}

\caption{Each model/method is evaluated through simulated conversations with all other models/methods. For Agency -- 0: \textit{low}, 1: \textit{medium}, 2: \textit{high} agency. For Intentionality (I), Motivation (M), Self-Efficacy (SE), and Self-Regulation (SR) -- 0: \textit{no expression}, 1: \textit{moderate expression}, 2: \textit{strong expression}. Numbers highlighted in \valingood{blue} and \valinbad{red} are significantly better and worse respectively than the overall mean ($p < 0.05$).}

\label{tab:dialogue-eval}
\vspace{-10pt}
\end{table}



\xhdr{GPT-4 demonstrates high Agency} Of the models tested, we find that GPT-4 demonstrates significantly higher Agency than others ($p < 0.05$). It particularly demonstrates the highest Intentionality which we found to have a strong correlation with Agency (Section~\ref{subsec:relationship-features}). Also, both GPT-4 and GPT-3 demonstrate significantly higher Self-Efficacy, indicating effectiveness in pursuing preferences and arguments ($p < 0.05$).

\begin{figure*}[t]
\centering
\vspace{-10pt}
\includegraphics[width=\textwidth]{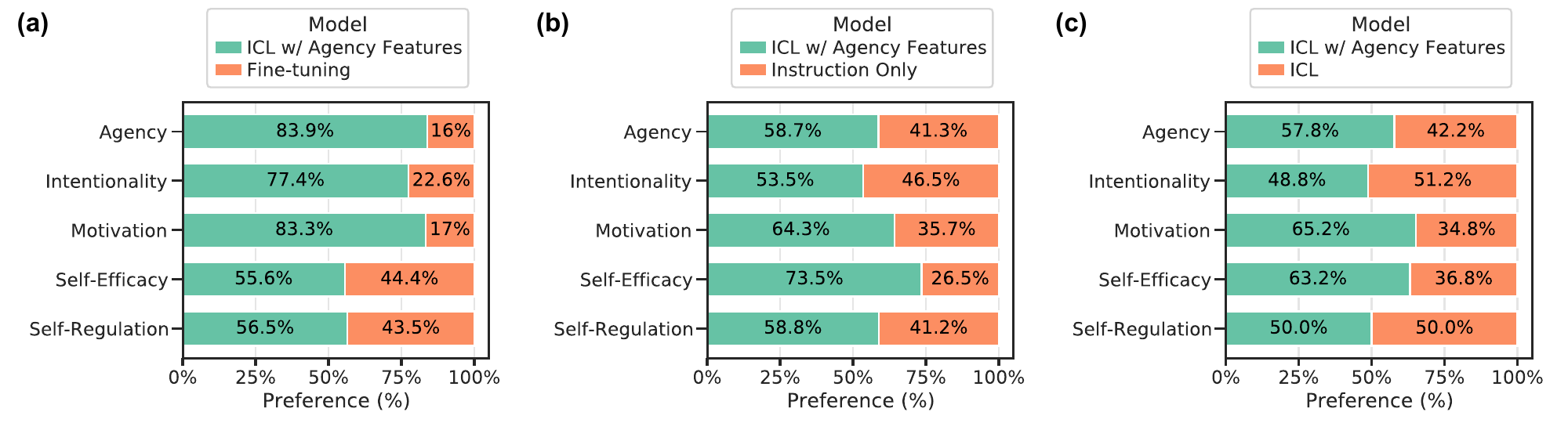}
\vspace{-15pt}
\caption{Human Evaluation Results. }
\label{human-eval}
\vspace{-10pt}
\end{figure*}

\xhdr{Llama2 demonstrates high Motivation, but low Self-Efficacy and Self-Regulation} We find that Llama2 variants demonstrate high Motivation, indicative of their reasoning capabilities that enable them to offer strong supportive evidence. However, they have lower Self-Efficacy and Self-Regulation indicating that it is relatively challenging to sustain their preferences and arguments, which may ultimately lead to lower agency. Guanaco similarly demonstrates significantly lower Self-Efficacy than other models ($p < 0.05$).

\xhdr{\editsc{Larger models demonstrate higher Intentionality, but lower Self-Efficacy}} Llama2 variants with more parameters have higher Intentionality, but lower Self-Efficacy. This suggests that while a larger model size can enhance the expression of preferences, it might not necessarily facilitate the sustained pursuit of those preferences and reasons over multiple conversational turns.

\subsection{Variation in Agency based on Finetuning/Prompting Methods}
\label{subsec:dialogue-methods}

We investigate the variations in Agency based on four different finetuning/prompting methods. We use a single model in this experiment.\footnote{We chose GPT-3 over GPT-4 because GPT-4 doesn't support fine-tuning, and GPT-3 offers the next best agency.}


\xhdr{Fine-tuning} We use the dataset collected by us (Section~\ref{sec:data}) to fine-tune GPT-3 (Appendix \ref{appendix:model-configurations}). 

\xhdr{Instruction Only} We prompt GPT-3 with the instruction used in Section~\ref{subsec:dialogue-models}. 

\xhdr{In-Context Learning (ICL)} We randomly retrieve $k$ conversational snippets from our dataset and construct demonstration examples. 


\xhdr{In-Context Learning w/ Agency Feature Examples (ICL-Agency)} We retrieve $k$ conversational snippets that score highly on our four Agency features 
and employ them as demonstration examples in a setup similar to the previous baseline.



Table~\ref{tab:dialogue-eval} shows the automatic evaluation results. The fine-tuned model struggles with this task. Qualitative analysis suggests that the generated responses from the fine-tuned model tend to be shorter, less natural, and less readable, potentially impacting its performance. In-Context Learning is better at expressing Intentionality and Motivation than the Instruction Only model, indicating that demonstration examples help. Finally, the highest value on all five metrics is achieved by In-Context Learning w/ Agency Feature Examples, highlighting the importance of incorporating examples related to these features in this task.

\subsection{Human Evaluation}
We evaluate the Agency of our best-performing method based on automatic evaluation, \textit{ICL-Agency}, with human interior designers (Figure~\ref{human-eval}). 

\xhdr{Procedure} We recruit 13 interior designers from UpWork (\href{https://www.upwork.com/}{upwork.com}). In each evaluation session, we ask them to interact with two randomly-ordered dialogue systems -- \textit{ICL-Agency} and one of the other three finetuning/prompting methods -- one at a time. They were provided with a room description and a chair design element (e.g., material). After their interaction, we asked them to choose the chatbot that had the (1) higher Agency, (2) higher Intentionality, (3) higher Motivation, (4) higher Self-Efficacy, and (5) higher Self-Regulation. \editsc{The designers conducted 231 comparative evaluations for a total of 231*2 = 462 interactions with LLM.\footnote{\editsc{We aimed to collect 20 evaluations per designer, but some dropped out before finishing all 20.}}}




\xhdr{Results} Consistent with the automatic evaluation results, ICL w/ Agency Features model is rated as having more Agency compared to other models and the Fine-tuning model is rated the worst. We do not observe significant differences in Intentionality between this model and the Instruction Only and In-Context Learning approaches. However, we find that this model is perceived as more effective in Motivation and Self-Efficacy, likely due to better access to relevant demonstration examples. 

\section{Further Related Work}
Previous dialogue research has studied personalized persuasive
dialogue systems \cite{wang2019persuasion}. Researchers have also built systems for negotiation tasks such as bargaining for goods \cite{he2018decoupling,joshidialograph} and strategy games like Diplomacy \cite{meta2022human}. Our work studies the broader concept of Agency and how dialogue systems may contribute to tasks through language. Research on creative AI has explored how collaboration b/w human and AI can be facilitated through dialogue in applications like collaborative drawing \cite{kim2019codraw} and facial editing \cite{jiang2021talk}. Here, we focus on the interior designing application as it presents significant complexity in terms of how Agency is shared. 

Agency has been studied in the context of undesirable biases in stories and narratives \cite{sap2017connotation} and how controllable revisions can be used to portray characters with more power and agency \cite{ma2020powertransformer}. In other domains such as games, researchers have created frameworks of Agency between players \cite{harrell2009agency,pickett2015npcagency,cole2018connecting,moallem2020review}. Our work develops a framework for measuring Agency in dialogue and explores how dialogue systems can be imbued with Agency. 
\section{Discussion and Conclusion}
\label{sec:discussion}
The idea of AI systems with Agency stems from the discourse surrounding the development of autonomous intelligent agents capable of mimicking human-like behavior and decision-making \cite{harrell2009agency,wen2022sense}. Agency drives how an agent contributes to a given task. 
In settings like games or AI-assisted teaching, AI may be the one guiding the task (e.g., as a non-character player). Also, in creative applications, engaging with a reactive AI without intention, motivation, and goals may be perceived as less meaningful.

\editsc{The ideal Agency of an agent would be defined by the task/application. Moreover, varying degrees of Agency might need to be manifested at different points in the interaction with a human. Learning how to best modulate the Agency based on the task and the ongoing human-LLM interaction forms an important future direction of work. Developing methods that effectively elicit and model task preferences for Agency and adapt LLMs based on the degrees to which they should actively contribute to the task, could be helpful in achieving this goal. Such methods could make use of the datasets and methods that we develop for assessing Agency levels of LLMs.}

The four features of Agency can be in conflict with each other, as well as with the Agency of the interlocutor. Thus, understanding how to detect and measure these features can help create agents who might converse more naturally and match the character of their human interlocutor. Importantly, our measurements of Agency and its features may be used to control the level of Agency in dialogue systems since different individuals may have different preferences on the desired amount of Agency across the four Agency features.

Although our dataset is focused on the domain of interior design, the Agency-related constructs that we introduce in this paper (e.g., \textit{Intentionality}) may be associated with domain-independent pragmatic features (e.g., ``\textit{I would prefer}'') and potentially permit adaptation to a variety of domains.



\section*{Ethics Statements}
This study was reviewed and approved by our Institutional Review Board. No demographic or Personal Identifiable Information was collected. Participants were paid \$20 per conversational session lasting no more than 30 minutes. Participants were based in US or Canada as reported through UpWork. Participant consent was obtained before starting the data collection.

Agency is a property with much potential to enhance collaborative interactions between human users and conversational agents. Nevertheless, full Agency may have unintended undesirable and potentially disruptive outcomes. In particular, the potential demonstrated in this work to control the degree of Agency may result in conversational agents being misapplied in disinformation campaigns or to manipulate for, e.g., financial gain.

\editsc{\section*{Acknowledgements}
We would like to thank Michael Xu for providing feedback on our study design and Alexandros Graikos for helping with ShapeNet and ShapeGlot datasets. We are also grateful to Ryen White and Nirupama Chandrasekaran for providing feedback through an initial pilot study. We also thank all the interior designers who contributed to data collection and human evaluation.}

\section*{Limitations}
Our experiments are restricted to the English language. We note that our dataset is focused on the domain of interior design. However, the Agency-related constructs we introduce in this paper, such as Intentionality, may also rely on domain-independent ``stylistic'' features (e.g., ``\textit{I would prefer}'') and could potentially be adapted to a variety of domains, which forms an interesting future direction of research. Also, our automatic measurements of Agency and its features are limited by the performance of the Agency prediction methods we tested. Future work may focus on designing more accurate automated Agency measurements.  

\bibliography{_references}

\begin{thebibliography}{42}
\expandafter\ifx\csname natexlab\endcsname\relax\def\natexlab#1{#1}\fi

\bibitem[{Achlioptas et~al.(2019)Achlioptas, Fan, Hawkins, Goodman, and Guibas}]{achlioptas2019shapeglot}
Panos Achlioptas, Judy Fan, Robert Hawkins, Noah Goodman, and Leonidas~J Guibas. 2019.
\newblock Shapeglot: Learning language for shape differentiation.
\newblock In \emph{ICCV}.

\bibitem[{Adiwardana et~al.(2020)Adiwardana, Luong, So, Hall, Fiedel, Thoppilan, Yang, Kulshreshtha, Nemade, Lu et~al.}]{adiwardana2020towards}
Daniel Adiwardana, Minh-Thang Luong, David~R So, Jamie Hall, Noah Fiedel, Romal Thoppilan, Zi~Yang, Apoorv Kulshreshtha, Gaurav Nemade, Yifeng Lu, et~al. 2020.
\newblock Towards a human-like open-domain chatbot.
\newblock \emph{arXiv preprint arXiv:2001.09977}.

\bibitem[{Bakhtin et~al.(2022)Bakhtin, Brown, Dinan, Farina, Flaherty, Fried, Goff, Gray, Hu et~al.}]{meta2022human}
Anton Bakhtin, Noam Brown, Emily Dinan, Gabriele Farina, Colin Flaherty, Daniel Fried, Andrew Goff, Jonathan Gray, Hengyuan Hu, et~al. 2022.
\newblock Human-level play in the game of diplomacy by combining language models with strategic reasoning.
\newblock \emph{Science}, 378(6624):1067--1074.

\bibitem[{Banaei et~al.(2017)Banaei, Ahmadi, and Yazdanfar}]{banaei2017application}
Maryam Banaei, Ali Ahmadi, and Abbas Yazdanfar. 2017.
\newblock Application of ai methods in the clustering of architecture interior forms.
\newblock \emph{Frontiers of Architectural Research}, 6(3):360--373.

\bibitem[{Bandura(1989)}]{bandura1989human}
Albert Bandura. 1989.
\newblock Human agency in social cognitive theory.
\newblock \emph{American psychologist}.

\bibitem[{Bandura(2000)}]{bandura2000exercise}
Albert Bandura. 2000.
\newblock Exercise of human agency through collective efficacy.
\newblock \emph{Current directions in psychological science}.

\bibitem[{Bandura(2001)}]{bandura2001social}
Albert Bandura. 2001.
\newblock Social cognitive theory: An agentic perspective.
\newblock \emph{Annual review of psychology}.

\bibitem[{Bates et~al.(1994)}]{bates1994role}
Joseph Bates et~al. 1994.
\newblock The role of emotion in believable agents.
\newblock \emph{Communications of the ACM}, 37(7):122--125.

\bibitem[{Binz and Schulz(2023)}]{binz2023using}
Marcel Binz and Eric Schulz. 2023.
\newblock Using cognitive psychology to understand gpt-3.
\newblock \emph{Proceedings of the National Academy of Sciences}, 120(6):e2218523120.

\bibitem[{Brown et~al.(2020)Brown, Mann, Ryder, Subbiah, Kaplan, Dhariwal, Neelakantan, Shyam, Sastry, Askell et~al.}]{brown2020language}
Tom Brown, Benjamin Mann, Nick Ryder, Melanie Subbiah, Jared~D Kaplan, Prafulla Dhariwal, Arvind Neelakantan, Pranav Shyam, Girish Sastry, Amanda Askell, et~al. 2020.
\newblock Language models are few-shot learners.
\newblock \emph{NeurIPS}.

\bibitem[{Chang et~al.(2015)Chang, Funkhouser, Guibas, Hanrahan, Huang, Li, Savarese, Savva, Song, Su et~al.}]{chang2015shapenet}
Angel~X Chang, Thomas Funkhouser, Leonidas Guibas, Pat Hanrahan, Qixing Huang, Zimo Li, Silvio Savarese, Manolis Savva, Shuran Song, Hao Su, et~al. 2015.
\newblock Shapenet: An information-rich 3d model repository.
\newblock \emph{arXiv preprint arXiv:1512.03012}.

\bibitem[{Chilton et~al.(2019)Chilton, Petridis, and Agrawala}]{chilton2019visiblends}
Lydia~B Chilton, Savvas Petridis, and Maneesh Agrawala. 2019.
\newblock Visiblends: A flexible workflow for visual blends.
\newblock In \emph{CHI}.

\bibitem[{Clark et~al.(2018)Clark, Ross, Tan, Ji, and Smith}]{clark2018creative}
Elizabeth Clark, Anne~Spencer Ross, Chenhao Tan, Yangfeng Ji, and Noah~A Smith. 2018.
\newblock Creative writing with a machine in the loop: Case studies on slogans and stories.
\newblock In \emph{IUI}.

\bibitem[{Code(2020)}]{code2020agency}
Jillianne Code. 2020.
\newblock Agency for learning: Intention, motivation, self-efficacy and self-regulation.
\newblock \emph{Frontiers in Genetics}.

\bibitem[{Cole(2018)}]{cole2018connecting}
Alayna Cole. 2018.
\newblock Connecting player and character agency in videogames.
\newblock \emph{Text}.

\bibitem[{Dettmers et~al.(2023)Dettmers, Pagnoni, Holtzman, and Zettlemoyer}]{dettmers2023qlora}
Tim Dettmers, Artidoro Pagnoni, Ari Holtzman, and Luke Zettlemoyer. 2023.
\newblock Qlora: Efficient finetuning of quantized llms.
\newblock \emph{arXiv preprint arXiv:2305.14314}.

\bibitem[{Emirbayer and Mische(1998)}]{emirbayer1998agency}
Mustafa Emirbayer and Ann Mische. 1998.
\newblock What is agency?
\newblock \emph{American journal of sociology}.

\bibitem[{Harrell and Zhu(2009)}]{harrell2009agency}
D~Fox Harrell and Jichen Zhu. 2009.
\newblock Agency play: Dimensions of agency for interactive narrative design.
\newblock In \emph{AAAI spring symposium: Intelligent narrative technologies II}.

\bibitem[{He et~al.(2018)He, Chen, Balakrishnan, and Liang}]{he2018decoupling}
He~He, Derek Chen, Anusha Balakrishnan, and Percy Liang. 2018.
\newblock Decoupling strategy and generation in negotiation dialogues.
\newblock In \emph{EMNLP}.

\bibitem[{Horton(2023)}]{horton2023large}
John~J Horton. 2023.
\newblock Large language models as simulated economic agents: What can we learn from homo silicus?
\newblock Technical report, National Bureau of Economic Research.

\bibitem[{Jiang et~al.(2021)Jiang, Huang, Pan, Loy, and Liu}]{jiang2021talk}
Yuming Jiang, Ziqi Huang, Xingang Pan, Chen~Change Loy, and Ziwei Liu. 2021.
\newblock Talk-to-edit: Fine-grained facial editing via dialog.
\newblock In \emph{ICCV}.

\bibitem[{Joshi et~al.(2021)Joshi, Balachandran, Vashishth, Black, and Tsvetkov}]{joshidialograph}
Rishabh Joshi, Vidhisha Balachandran, Shikhar Vashishth, Alan Black, and Yulia Tsvetkov. 2021.
\newblock Dialograph: Incorporating interpretable strategy-graph networks into negotiation dialogues.
\newblock In \emph{ICLR}.

\bibitem[{Kant(1951)}]{kant1951critique}
Immanuel Kant. 1951.
\newblock Critique of judgment, trans. jh bernard.
\newblock \emph{New York: Hafner}.

\bibitem[{Kim et~al.(2019)Kim, Kitaev, Chen, Rohrbach, Zhang, Tian, Batra, and Parikh}]{kim2019codraw}
Jin-Hwa Kim, Nikita Kitaev, Xinlei Chen, Marcus Rohrbach, Byoung-Tak Zhang, Yuandong Tian, Dhruv Batra, and Devi Parikh. 2019.
\newblock Codraw: Collaborative drawing as a testbed for grounded goal-driven communication.
\newblock In \emph{ACL}.

\bibitem[{Koo et~al.(2022)Koo, Huang, Achlioptas, Guibas, and Sung}]{koo2022partglot}
Juil Koo, Ian Huang, Panos Achlioptas, Leonidas~J Guibas, and Minhyuk Sung. 2022.
\newblock Partglot: Learning shape part segmentation from language reference games.
\newblock In \emph{ICCV}.

\bibitem[{Liu et~al.(2019)Liu, Ott, Goyal, Du, Joshi, Chen, Levy, Lewis, Zettlemoyer, and Stoyanov}]{liu2019roberta}
Yinhan Liu, Myle Ott, Naman Goyal, Jingfei Du, Mandar Joshi, Danqi Chen, Omer Levy, Mike Lewis, Luke Zettlemoyer, and Veselin Stoyanov. 2019.
\newblock Roberta: A robustly optimized bert pretraining approach.
\newblock \emph{arXiv preprint arXiv:1907.11692}.

\bibitem[{Locke(1978)}]{locke1978two}
John Locke. 1978.
\newblock Two treatises of government.
\newblock \emph{New York: E. P. Dutton}.

\bibitem[{Ma et~al.(2020)Ma, Sap, Rashkin, and Choi}]{ma2020powertransformer}
Xinyao Ma, Maarten Sap, Hannah Rashkin, and Yejin Choi. 2020.
\newblock Powertransformer: Unsupervised controllable revision for biased language correction.
\newblock In \emph{EMNLP}.

\bibitem[{Moallem and Raffe(2020)}]{moallem2020review}
Jonathan~D Moallem and William~L Raffe. 2020.
\newblock A review of agency architectures in interactive drama systems.
\newblock In \emph{2020 IEEE Conference on Games (CoG)}, pages 305--311. IEEE.

\bibitem[{Oh et~al.(2018)Oh, Song, Choi, Kim, Lee, and Suh}]{oh2018lead}
Changhoon Oh, Jungwoo Song, Jinhan Choi, Seonghyeon Kim, Sungwoo Lee, and Bongwon Suh. 2018.
\newblock I lead, you help but only with enough details: Understanding user experience of co-creation with artificial intelligence.
\newblock In \emph{CHI}.

\bibitem[{OpenAI(2023)}]{OpenAI2023GPT4TR}
OpenAI. 2023.
\newblock \href {https://api.semanticscholar.org/CorpusID:257532815} {Gpt-4 technical report}.
\newblock \emph{ArXiv}, abs/2303.08774.

\bibitem[{Park et~al.(2023)Park, O'Brien, Cai, Morris, Liang, and Bernstein}]{park2023generative}
Joon~Sung Park, Joseph~C O'Brien, Carrie~J Cai, Meredith~Ringel Morris, Percy Liang, and Michael~S Bernstein. 2023.
\newblock Generative agents: Interactive simulacra of human behavior.
\newblock \emph{arXiv preprint arXiv:2304.03442}.

\bibitem[{Pickett et~al.(2015)Pickett, Fowler, and Khosmood}]{pickett2015npcagency}
Grant Pickett, Allan Fowler, and Foaad Khosmood. 2015.
\newblock Npcagency: conversational npc generation.
\newblock In \emph{Proceedings of the 10th International Conference on the Foundations of Digital Games}.

\bibitem[{Riedl and Bulitko(2012)}]{riedl2012interactive}
Mark Riedl and Vadim Bulitko. 2012.
\newblock Interactive narrative: A novel application of artificial intelligence for computer games.
\newblock In \emph{Proceedings of the AAAI Conference on Artificial Intelligence}, volume~26, pages 2160--2165.

\bibitem[{Roller et~al.(2021)Roller, Dinan, Goyal, Ju, Williamson, Liu, Xu, Ott, Smith, Boureau et~al.}]{roller2021recipes}
Stephen Roller, Emily Dinan, Naman Goyal, Da~Ju, Mary Williamson, Yinhan Liu, Jing Xu, Myle Ott, Eric~Michael Smith, Y-Lan Boureau, et~al. 2021.
\newblock Recipes for building an open-domain chatbot.
\newblock In \emph{ACL}.

\bibitem[{Sap et~al.(2017)Sap, Prasettio, Holtzman, Rashkin, and Choi}]{sap2017connotation}
Maarten Sap, Marcella~Cindy Prasettio, Ari Holtzman, Hannah Rashkin, and Yejin Choi. 2017.
\newblock Connotation frames of power and agency in modern films.
\newblock In \emph{EMNLP}.

\bibitem[{Touvron et~al.(2023)Touvron, Martin, Stone, Albert, Almahairi, Babaei, Bashlykov, Batra, Bhargava, Bhosale et~al.}]{touvron2023llama}
Hugo Touvron, Louis Martin, Kevin Stone, Peter Albert, Amjad Almahairi, Yasmine Babaei, Nikolay Bashlykov, Soumya Batra, Prajjwal Bhargava, Shruti Bhosale, et~al. 2023.
\newblock Llama 2: Open foundation and fine-tuned chat models.
\newblock \emph{arXiv preprint arXiv:2307.09288}.

\bibitem[{Volum et~al.(2022)Volum, Rao, Xu, DesGarennes, Brockett, Van~Durme, Deng, Malhotra, and Dolan}]{volum2022craft}
Ryan Volum, Sudha Rao, Michael Xu, Gabriel DesGarennes, Chris Brockett, Benjamin Van~Durme, Olivia Deng, Akanksha Malhotra, and William~B Dolan. 2022.
\newblock Craft an iron sword: Dynamically generating interactive game characters by prompting large language models tuned on code.
\newblock In \emph{Proceedings of the 3rd Wordplay: When Language Meets Games Workshop (Wordplay 2022)}, pages 25--43.

\bibitem[{Wang et~al.(2019)Wang, Shi, Kim, Oh, Yang, Zhang, and Yu}]{wang2019persuasion}
Xuewei Wang, Weiyan Shi, Richard Kim, Yoojung Oh, Sijia Yang, Jingwen Zhang, and Zhou Yu. 2019.
\newblock Persuasion for good: Towards a personalized persuasive dialogue system for social good.
\newblock In \emph{ACL}.

\bibitem[{Wei et~al.(2022)Wei, Wang, Schuurmans, Bosma, Xia, Chi, Le, Zhou et~al.}]{weichain}
Jason Wei, Xuezhi Wang, Dale Schuurmans, Maarten Bosma, Fei Xia, Ed~H Chi, Quoc~V Le, Denny Zhou, et~al. 2022.
\newblock Chain-of-thought prompting elicits reasoning in large language models.
\newblock In \emph{NeurIPS}.

\bibitem[{Wen and Imamizu(2022)}]{wen2022sense}
Wen Wen and Hiroshi Imamizu. 2022.
\newblock The sense of agency in perception, behaviour and human--machine interactions.
\newblock \emph{Nature Reviews Psychology}, 1(4):211--222.

\bibitem[{Ziems et~al.(2023)Ziems, Held, Shaikh, Chen, Zhang, and Yang}]{ziems2023can}
Caleb Ziems, William Held, Omar Shaikh, Jiaao Chen, Zhehao Zhang, and Diyi Yang. 2023.
\newblock Can large language models transform computational social science?
\newblock \emph{arXiv preprint arXiv:2305.03514}.

\end{thebibliography}

\appendix
\clearpage
\clearpage

\section{Dataset Statistics}
\label{appendix:data-stats}
\begin{table}[h!]
\centering
\resizebox{1.0\columnwidth}{!}{
\def\arraystretch{1.2}
\begin{tabular}{lcccc}
\toprule
\textbf{Feature} & \textbf{N/A} & \textbf{No} & \textbf{Moderate} & \textbf{Strong} \\
\toprule
Intentionality & -- & 194 & 175 & 539 \\
Motivation & -- & 474 & 158 & 276 \\
Self-Efficacy & 770 & 63 & 46 & 29 \\
Self-Regulation & 764 & 25 & 61 & 58 \\
\bottomrule
\end{tabular}
}
\caption{Statistics of the annotated conversation snippets. N/A indicates not applicable. We annotate Self-Efficacy as N/A if a designer never indicated a preference or did not need to pursue their preference (e.g., because the other designer did not argue against it). We annotate Self-Regulation as N/A if a designer Never indicated a preference or did not need to change their preference (e.g., because the other designer did not argue against it).}
\label{tab:data-stats}
\end{table}

\begin{table}[h!]
\centering
\def\arraystretch{1.2}
\begin{tabular}{lcccc}
\toprule
\textbf{} & \textbf{Low} & \textbf{Medium} & \textbf{High} \\
\toprule
Agency & 308 & 292 & 308  \\
\bottomrule
\end{tabular}
\caption{Agency distribution of the conversation snippets.}
\label{tab:data-stats-agency}
\end{table}

\xhdr{Other Statistics} The conversations b/w interior designers in our dataset have 41.67 turns on average. The extracted conversation snippets have 4.21 turns on average. We find an average pairwise agreement of 71.36\% for Intentionality, 70.70\% for Motivation, 85.21\% for Self-Efficacy, and 81.09\% for Self-Regulation.

\section{Model Details}
\label{appendix:model-configurations}
We use \texttt{text-davinci-003} for all of our GPT-3 models. For Agency measurement models (Section~\ref{sec:prediction}), we sample the highest probable next tokens by setting the temperature value to $0$ (determinstic sampling). For dialogue generation models (Section~\ref{sec:dialogue}), we use top-p sampling with $p=0.6$. For in-context learning methods, we experimented with $k=5,10,15$, and $20$ and found $k=10$ to be the most effective based on a qualitative assessment of 10 examples.

\xhdr{GPT-3 (Q/A)} We frame our measurement tasks as conversational question-answering. For a given conversational snippet, we ask GPT-3 \cite{brown2020language} to answer the questions related to each of the five subtasks (same questions as asked during data collection (Section~\ref{subsec:annotated-data})). We present $k=10$ demonstration examples, randomly sampled from our dataset (different examples for each of the five subtasks; Appendix~\ref{apendix:gpt-3-q-a}).

\xhdr{GPT-3 (CoT) and GPT-4 (CoT)} We use chain-of-thought (CoT) prompting \cite{weichain} to reason about conversational snippets. We use $k=10$ demonstration examples, randomly sampled from our dataset and manually write chain-of-thought prompts for each of the five subtasks

\xhdr{Fine-tuning details} Since our goal is to simulate a dialogue agent with high Agency, for each conversational snippet, we label the designer who influenced the design (who had a higher agency) as ``AI'' and the other designer (who had a lower agency) as ``Human''. We fine-tune GPT-3 to generate AI utterances given all previous utterances in a conversational snippet and the instruction prompt developed for the Instruction Only baseline.

\section{Linguistic Attributes Measurement}
\label{appendix:linguistic-attributes}
We compare the tentativeness, self-focus, reasoning, and persuasion of the designers using the following prompts. We randomly assign the names of \textit{Tom} and \textit{Harry} to the two designers.

\xhdr{Tentativeness} \textit{Your job is to assess tentativeness in a conversation between Tom and Harry about designing chairs. A tentaitve person will not be confident about their arguments.}

\xhdr{Self-Focus} \textit{Your job is to assess self-focusedness in a conversation between Tom and Harry about designing chairs. A self-focused person will be more focused on their own arguments than the other person's arguments.}

\xhdr{Reasoning} \textit{Your job is to assess reasoning strength in a conversation between Tom and Harry about designing chairs. A person with strong reasoning will have strong arguments.}

\xhdr{Persuasion} \textit{Your job is to assess persuasion in a conversation between Tom and Harry about designing chairs. A persuasive person will be able to convince the other person about their arguments.}

\section{Human Evaluation Details}
We asked three evaluators to choose the chatbot that \textbf{(1)} had more influence over the final design (Agency); \textbf{(2)} was better able to express its design preference (Intentionality); \textbf{(3)} was better able to motivate their design preference (Motivation); \textbf{(4)} pursued their design preferences for a greater number of conversational turns (Self-Efficacy); \textbf{(5)} was better able to self-adjust their preference (Self-Regulation).

\section{Why We Chose Collaborative Interior Designing as Our Testbed?}
\label{appendix:interior-design}
Here, we propose a \textbf{dialogue-based collaborative interior design task} as a testbed. In this task, given a room setting, the goal is to discuss how to design the interiors of the room.

We note that an interior design task can be broad and may involve a wide range of complex components (e.g., color palette, furniture, accessories) as well as a series of steps to be followed. Furthermore, due to a real-world room context, the task must be grounded with both vision and language components with an understanding of how three-dimensional objects in a room (e.g., chairs, tables, plants, decor items) must be designed. 

Here, we build upon previous work on richly-annotated, large-scale datasets of 3D objects like ShapeNet \cite{chang2015shapenet} and subsequent works on understanding how fine-grained differences between objects are expressed in language like ShapeGlot \cite{achlioptas2019shapeglot} and PartGlot \cite{koo2022partglot}. Both ShapeGlot and PartGlot datasets provide us with richly annotated datasets of chairs. Therefore, we narrow down the scope of our task and specifically focus on \textit{furnishing a room with a chair}. In this task, a human and an AI are provided with a room layout and asked to collaboratively come up with a design of a chair to be placed in the room through text-based interaction. 


\section{Analysis of Agency Features}

\begin{table}[h!]
\centering
\def\arraystretch{1.2}
\begin{tabular}{lc}
\toprule
\textbf{Agency Feature} & \textbf{Coefficient} \\
\toprule
Intentionality & 0.1435* \\
Motivation &  0.0235 \\
Self-Efficacy & 0.0384 \\
Self-Regulation & -0.1224* \\
\bottomrule
\end{tabular}
\caption{Coefficients for predicting agency in conversations using a mixed-effect linear regression model. *$p < 0.05$}
\label{tab:linear-model}
\end{table}

\section{Task 1: Demonstration examples}
\subsection{GPT-3 (Q/A)}
\label{apendix:gpt-3-q-a}
For the GPT-3 (Q/A) model, we present examples to GPT-3 in the following format:

\begin{quote}
    \textbf{Designer:} I think a black wooden frame or black metal legs (to match the bed frame) would work. \\
    \textbf{Other Designer:} I like the black metal legs.  What about hairpin legs? \\
    \textbf{Designer:} Or maybe brass legs would be better. Hairpin legs would work fine, but would the rest of the frame be the black wood? \\
    \textbf{Other Designer:} If we did brass tapered metal legs it would tie well with the black wood. \\
    \textbf{Designer:} I think that would look better.
    Other Designer: Agreed \\
    
    \textbf{Who influenced the design element being discussed?:} Other Designer \\
\end{quote}

\subsection{GPT-3 (CoT)}
\label{apendix:gpt-3-chain-of-thought}

For the GPT-3 (CoT) model, we present examples to GPT-3 in the following format:

\begin{quote}
    \textbf{Designer:} I think a black wooden frame or black metal legs (to match the bed frame) would work. \\
    \textbf{Other Designer:} I like the black metal legs.  What about hairpin legs? \\
    \textbf{Designer:} Or maybe brass legs would be better. Hairpin legs would work fine, but would the rest of the frame be the black wood? \\
    \textbf{Other Designer:} If we did brass tapered metal legs it would tie well with the black wood. \\
    \textbf{Designer:} I think that would look better.
    Other Designer: Agreed \\

    \textbf{TL;dr} Brass tapered metal legs were agreed upon. This was initially proposed by the Other Designer. 
\end{quote}

\section{Reproducibility}

 We release the code and datasets developed in this paper at  \href{https://github.com/microsoft/agency-dialogue}{github.com/microsoft/agency-dialogue}. 
 
 The use of existing artifacts conformed to their intended use. We used the OpenAI library for GPT-3 and GPT-4 based models. We used A100 GPUs to perform inference on Llama2 and Guanaco. We use the scipy and statsmodel libraries for statistical tests in this paper. 


 \clearpage

 \section{Human-Human Conversational Data Collection Instructions}
\label{appendix:data-collection-instructions}
\begin{figure}[hbt!]
    \caption{Instructions shown to the interior designers during the human-human conversational data collection. Continued on the next page (1/3).}
    \centering
         \includegraphics[width=\textwidth]{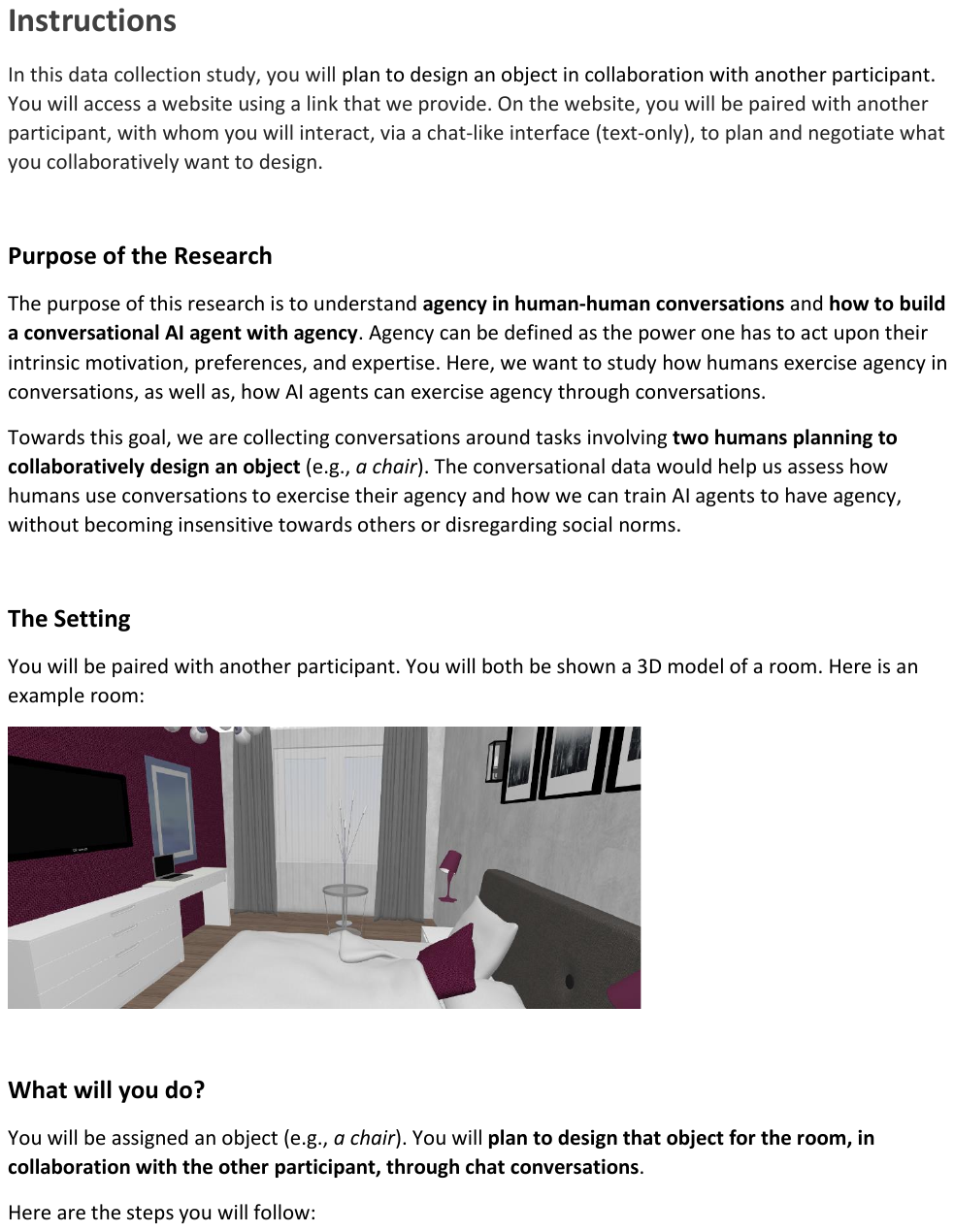}
    \label{supp:fig:data-instructions-1}
\end{figure}

\begin{figure*}[hbt!]
    \caption{Instructions shown to the interior designers during the human-human conversational data collection. Continued on the next page (2/3).}
    \centering
         \includegraphics[width=\textwidth]{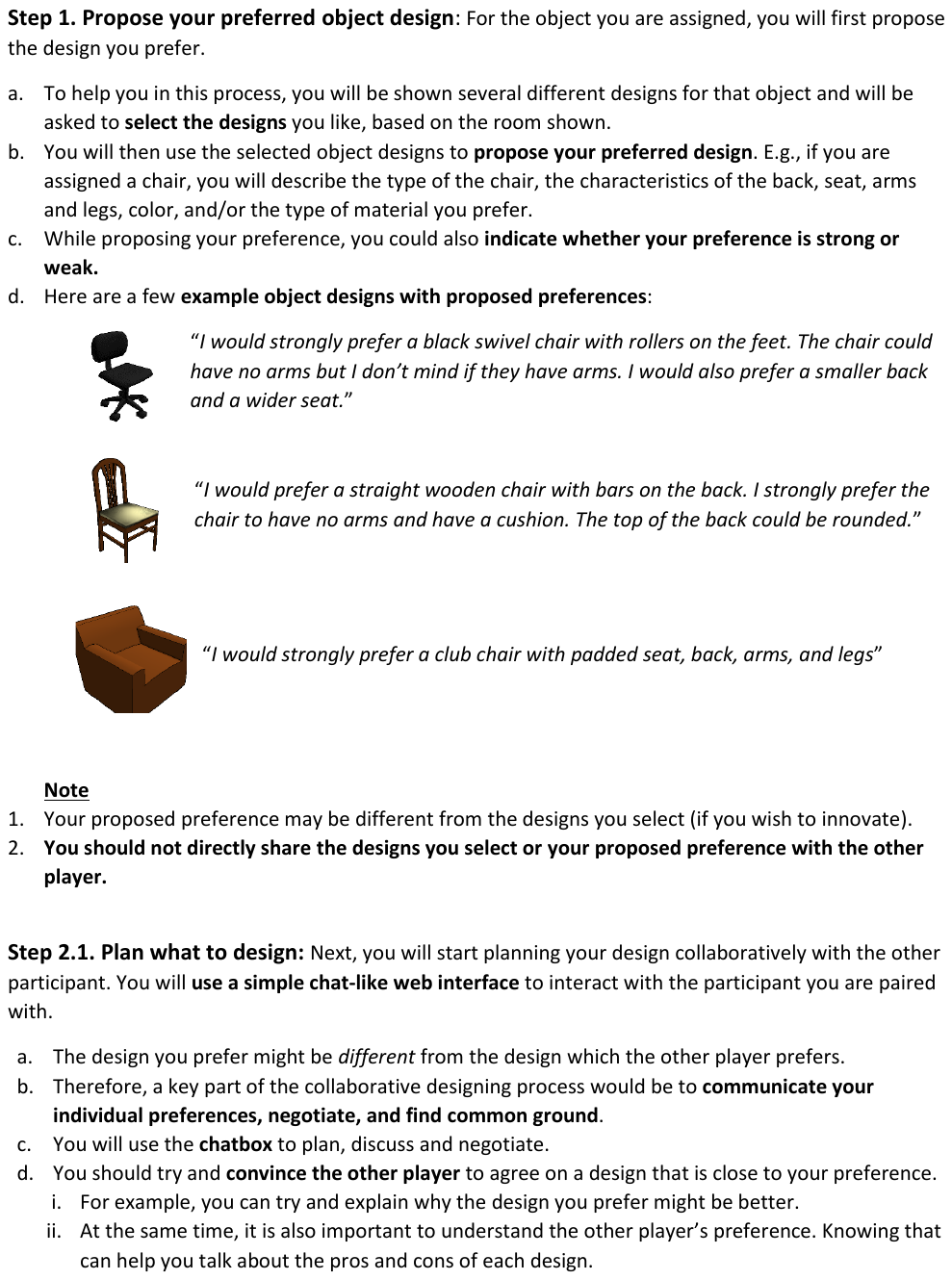}
    \label{supp:fig:data-instructions-2}
\end{figure*}

\begin{figure*}[hbt!]
    \caption{Instructions shown to the interior designers during the human-human conversational data collection (3/3).}
    \centering
         \includegraphics[width=\textwidth]{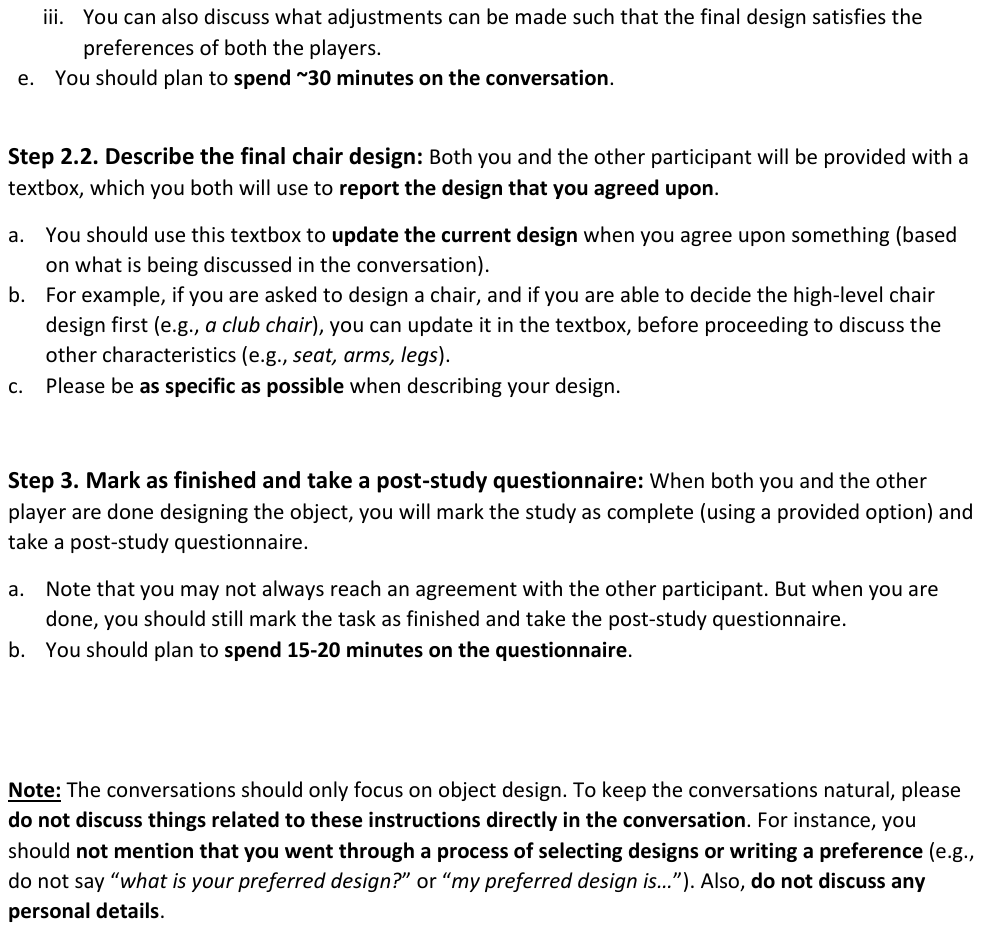}
    \label{supp:fig:data-instructions-3}
\end{figure*}

\begin{table*}[h!]
\centering
\resizebox{0.95\linewidth}{!}{
\def\arraystretch{1.2}
\begin{tabular}{lp{14cm}}
\toprule
\textbf{Designer} & \textbf{Utterance} \\
\toprule
Designer 1: & How about a desk chair for this area? \\
Designer 2: & There seems to be many possibilities for this space, would you agree? Yet I agree that some kind of chair for the desk is needed. \\
Designer 1: & The room has very clean lines with an Asian theme \\
Designer 2: & I think we need to support the minimalist lines of the overall space design. Not something too over-stuffed. Something with a contemporary feel. \\
Designer 1: & So maybe a more contemporary style of desk chair. \\
Designer 1: & Great minds! \\
Designer 1: & How do you feel about a tall back with tilt swivel and adjustable \\
Designer 2: & I believe so. Maybe one that is comfortable for sure - but not too closed in. There is the lovely background to consider. We don't want to block that. \\
Designer 1: & If not too tall, then maybe something mid back height? \\
Designer 2: & I think the height of the back should be carefully scaled - supportive but not so high that it obscures what is behind too much. \\
Designer 1: & Or shoulder height for support \\
Designer 1: & With arm support \\
Designer 2: & Agreed on shoulder height. Swiveling is good - also moving -like on casters may provide flexibility. \\
Designer 1: & Definitely casters \\
Designer 2: & I am concerned about tilting back since we do have some fragile decorative elements behind. \\
Designer 1: & Ok, so far... shoulder height desk chair with adjustable height, casters and arm rests \\
Designer 2: & I do agree that arm support is essential, especially if one is to feel comfortable while working. It  feels like this might be a consult room of sorts - so allowing the person to sit back in a more relaxed posture - resting arms off the table is good. \\
Designer 1: & Some tilts can be regulated and locked into place... not necessarily a full recline \\
Designer 1: & Perfect \\
Designer 2: & The materiality of the chair is something to consider. I see a lot of wood and timber detailing. It might be nice to have the chair upholsterable - perhaps a nice leather back that would be shaped to lightly massage the back? \\
Designer 1: & Agree \\
Designer 1: & the leather would be a nice look in there \\
Designer 2: & Something that seems pillowy or wavy, but in a very restrained, minimalist sort of way \\
Designer 1: & Black would match the ottomans but a soft buttery cream/ ivory would add a soothing neutral to the aesthetic \\
Designer 2: & With the darker wood in the room and the leather chair - an accent material on the armrests might be nice to offsett - say a brushed steel or aluminum finish? \\
Designer 1: & I've seen the vertical channeling on a desk chair that is very classy looking \\
Designer 1: & The brushed steel frame would look nice in this room. I think wood would be a bit much. \\
Designer 2: & I think classic modern which always took a lot of inspiration from japanese design. The buttery cream is a lovely idea. Will provide a bright focal point and it will align with the colors of the fan. \\
Designer 1: & I think we have our chair! \\
\bottomrule
\end{tabular}
}
\caption{Example Human-Human Conversation in Our Dataset.}
\label{appendix:tab:example-1}
\end{table*}

\clearpage

\section{Human Evaluation Experiment Instructions}

\label{appendix:human-eval-instructions}
\begin{figure}[hbt!]
    \caption{Instructions shown to the interior designers during the human evaluation experiment. Continued on the next page (1/2).}
    \centering
         \includegraphics[width=\textwidth]{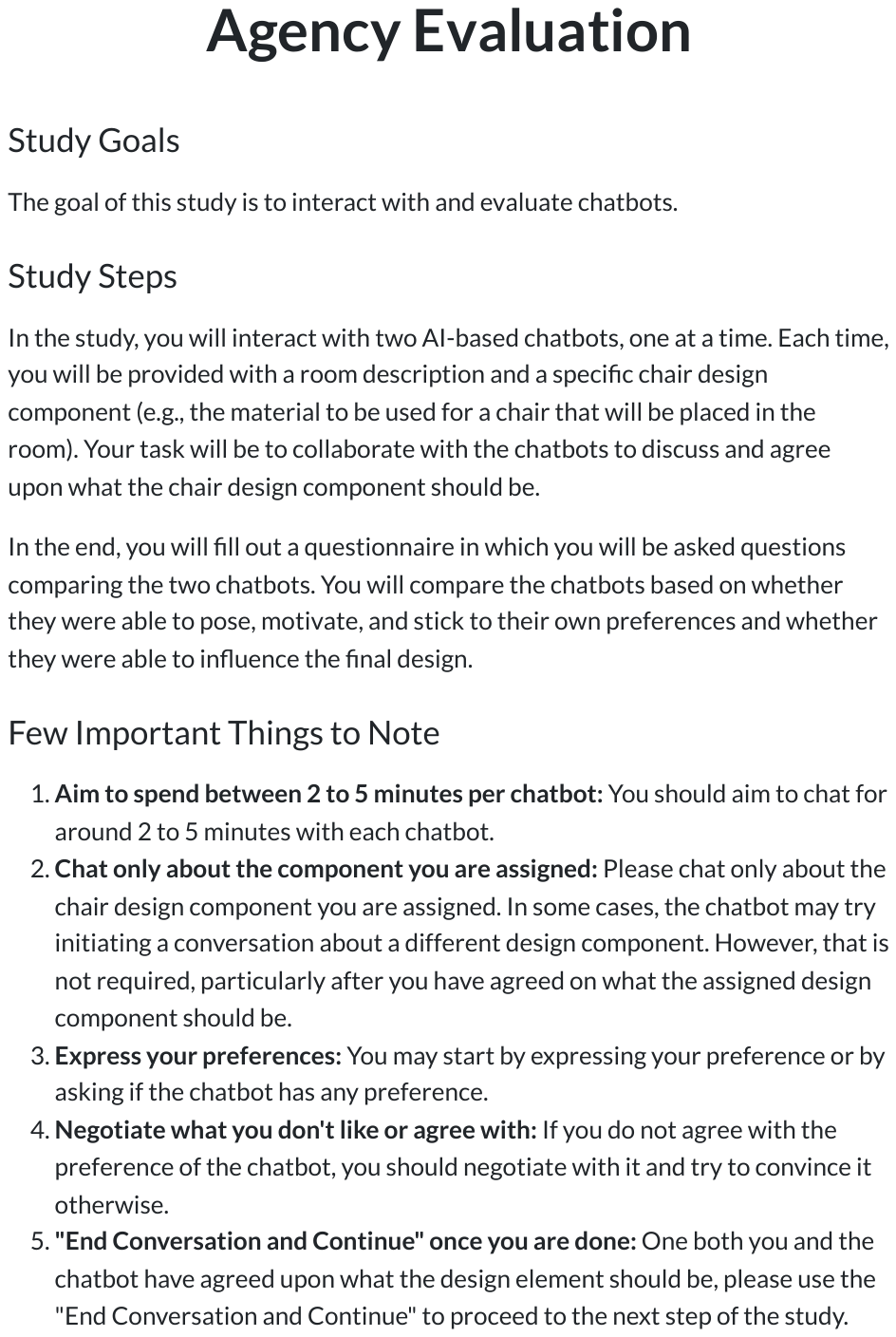}
    \label{supp:fig:human-eval-instructions-1}
\end{figure}

\begin{figure*}[hbt!]
    \caption{Instructions shown to the interior designers during the human evaluation experiment (2/2).}
    \centering
         \includegraphics[width=\textwidth]{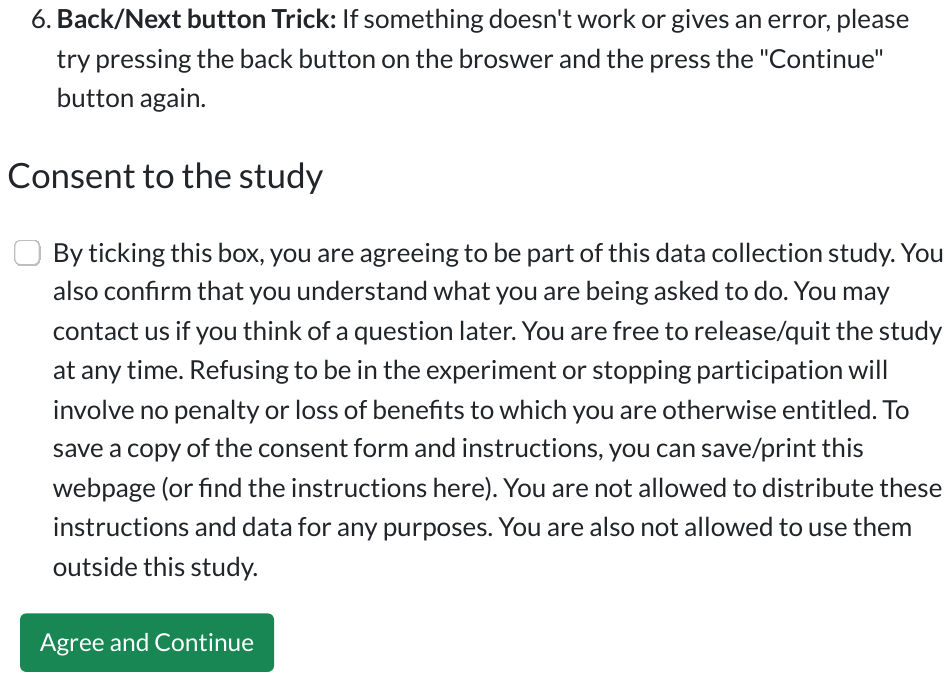}
    \label{supp:fig:human-eval-instructions-2}
\end{figure*}

\end{document}